\documentclass[10pt,journal,cspaper,compsoc]{IEEEtran}
\ifCLASSOPTIONcompsoc
\else
\fi

\ifCLASSINFOpdf
   \usepackage[pdftex]{graphicx}
\else
\fi

\usepackage{wrapfig}

\usepackage{cite}
\usepackage{url}

\usepackage{amsmath}

\usepackage{color} 
\usepackage{multirow,times}

\usepackage{epstopdf} 

\mathchardef\mhyphen="2D

\usepackage{maplestd2e}

\usepackage{hyperref} 

\usepackage{algorithmic}

\begin{document}

\bstctlcite{IEEEexample:BSTcontrol}

%
\title{Reflection Invariant and Symmetry Detection}

\author{Erbo Li and Hua Li
\IEEEcompsocitemizethanks{
\IEEEcompsocthanksitem E. Li is with EON Reality Inc, Irvine, CA 92618. Email: sophialiuli@gmail.com
\IEEEcompsocthanksitem H. Li is with Key lab of Intelligent Information Processing, Institute of Computing Technology, Chinese Academy of Sciences, 100190 Beijing; University of Chinese Academy of Sciences, 100049 Beijing. 
E-mail: lihua@ict.ac.cn
}
}


\markboth{}%
{Shell \MakeLowercase{\textit{Erbo Li et al.}}: Isomorphism between Differential and Moment Invariants under Affine Transform}

\IEEEcompsoctitleabstractindextext{%
\begin{abstract}

Symmetry detection and discrimination are of fundamental meaning in science,
technology, and engineering. This paper introduces reflection invariants and defines
the directional moments(DMs) to detect symmetry for shape analysis and object recognition.
And it demonstrates that detection of reflection symmetry can be done in a simple
way by solving a trigonometric system derived from the DMs, and
discrimination of reflection symmetry can be achieved by application of the reflection
invariants in 2D and 3D. Rotation symmetry can also be determined based on that. Also, if none of reflection invariants is equal to zero, then there is no symmetry. 
And the experiments in 2D and 3D show that all the reflection lines or planes can be deterministically
found using DMs up to order six. This result can be used to simplify the efforts of symmetry detection in research areas,such as protein structure, model retrieval, reverse engineering, and machine vision
etc.


\end{abstract}

\begin{IEEEkeywords}
symmetry detection, shape analysis, object recognition, directional moment, moment
invariant, isometry, congruent, reflection, chirality, rotation
\end{IEEEkeywords}}

\maketitle

\IEEEdisplaynotcompsoctitleabstractindextext

%
\IEEEpeerreviewmaketitle

\section{Introduction}
\label{sec:intro}

\begin{figure}[thb]
  \centering
  \includegraphics[width=0.99\linewidth]{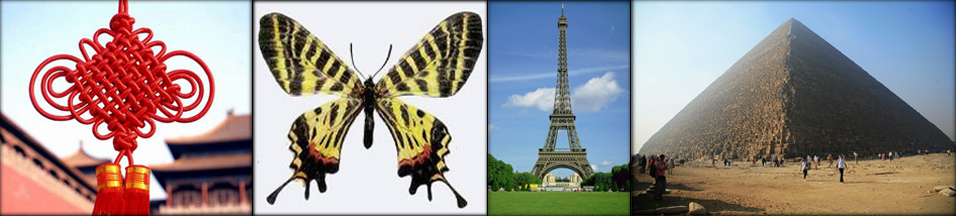}
  \caption{\label{fig:fig1}
  Symmetry is everywhere, where the pictures from left to right are Chinese knot, butterfly,
Eiffel Tower and Pyramid (The pictures shown here are from the Internet).  }
\end{figure}

The essence of geometric symmetry is self-evident, which can be found everywhere
in nature and social lives, as shown in Figure~\ref{fig:fig1}. It is true that we are living in a
symmetric world. Pursuing the explanation of symmetry will provide better
understanding to the surrounding world and the civilization inherited from history.
And in the art of painting, sculpture, and architecture etc. there are millions of
examples showing the existence of symmetry.

Basically, there are two kinds of symmetry: reflection and rotation. The former is
mirror symmetry and the later means that a shape will completely coincide with itself
after a rotation with some fixed angle about an axis. The number of the possible
rotations contained in a whole circle is called fold, whose parity leads to even or odd
symmetries. For example, the first two images in Figure~\ref{fig:fig1} are reflection symmetries
with a vertical line passing through their centers, and the last two are rotation
symmetries with four folds and therefore even symmetries.

Measuring and detecting symmetry in shape has a long history and have been paid
more interest till recently. Kazhdan et al. \cite{Kazhdan2003} developed a continuous measure and
discussed the properties of the reflective symmetry descriptor, which was expanded to
3D by \cite{Podolak2006} and was augmented in spatial distribution of the object's
asymmetry by \cite{Rustamov2007} . For symmetry discrimination \cite{Zabrodsky1995} defined
a symmetry distance of shapes. For geometric similarity or identity, Hu gave the
seven moment invariants and named the last moment invariant as skew invariant \cite{Hu1962},
which was actually useful for distinguishing "mirror images." Flusser et al. found a
new way to generate Hu's seven moment invariants \cite{Hu1962} \cite{Flusser2000} \cite{Gonzalez2007} \cite{Flusser2017} and reported some new
invariants to solve reflection problems in 2D. \cite{Teague1980} introduced complex
moment invariants, pointed some invariants changed sign under reflection and called
them pseudoinvariants, which was also adopted by \cite{Flusser2017}. \cite{Xu2008}
proposed an idea of geometric invariant cores, by which any order or degree of
moment invariants can be gained in different dimensions or manifolds by calculating
multiple integrals. \cite{Hattne2011} reported a moment invariant for 3D
reflection or chirality. \cite{Li2017} recently summed up the way of defining geometric
moment invariants and put forward two generating functions, called ShapeDNA,
based on which all the moment invariants can be obtained and provides a new viewangle to geometric inner structure and shape analysis.

In symmetry detection, there are various algorithms, such as segmentation by
searching each reflective symmetry plane \cite{Podolak2006} , defining general moment by cross
product of vectors \cite{Martinet2006}, detection by generalized complex moment \cite{Shen1999} , as well as other
detecting methods \cite{Hel-Or1991} \cite{Loy2006} \cite{Mitra2006} \cite{Tagliasacchi2009} \cite{Sun1997}.

Till now, the symmetry detection and discrimination are still difficult and
time-consuming. Two important progresses were by \cite{Shen1999} for 2D and
\cite{Martinet2006} for 3D. The former exploited properties of Fourier transform and the
later gave a way to accurately detect possible position of 3D symmetry, although it
needs a solution of complicated spherical harmonic expression of trigonometric
function.

Here we propose both the directional moment for symmetry detection and the
reflection invariant for symmetry discrimination. The combination of them provides a
powerful tool for the solution of geometric symmetry. Symmetry detection can be
done in a simple way by solving a trigonometric system derived from the directional
moment, and symmetry discrimination can be achieved by application of the
reflection invariant directly in 2D or 3D. Reflection symmetry is solved accurately
and deterministically. Rotation symmetry can also be dealt with. This work is direct
extension and complement of \cite{Martinet2006} and \cite{Shen1999} and unifies the solution for different
dimensions.

In this paper, we will first discuss the symmetry discrimination in Section 2, give the
definition of moment and classification of moment invariants, and then in Section 3
define the directional moment. The algorithms and experiments are given in Section 4.
Some discussions and conclusions are listed in Section 5 and 6 respectively.

\section{Symmetry discrimination}
\label{sec:2}

\subsection{Definition of moments}

The definition of moment in 2D has the following format

\textbf{Definition 1: moment}

An order (m+n) moment of a shape D is

\begin{equation}
\label{eq:1}
\mapleinline{inert}{2d}{M[mn] = int(int(x^m*y^n*rho(x, y), x), y)}{\[\displaystyle M_{{{\it mn}}}=\int \!\!\!\int \!{x}^{m}{y}^{n}\rho \left( x,y \right) {dx}\,{dy}\]}
\end{equation}

where $m, n$ are positive integers; $ \rho \left( x,y \right) $ is a density function; the integral is defined on shape D.

\textbf{Definition 2: central movement}

The central movement is

\begin{equation}
\label{eq:2}
\mu _{{\it mn} } = \int \!\!\! \int \left(x - x' \right)^{m } \left(y - y' \right)^{n }\rho \left(x ,y \right) dx dy
\end{equation}

where $x'$ and $y'$ are the coordinates of the mass center or centroid,

\begin{equation}
\label{eq:3}
x' = \frac {M_{{10}}}{M_{{00}}}, y' = \frac {M_{{01}}}{M_{{00}}}
\end{equation}

The central moment is used to define translation invariants and makes their expressions more concisely.

The definitions expressed in Eq.~\eqref{eq:1} and Eq.~\eqref{eq:2} can be easily extended to higher dimensions or on different manifolds.

Moments have a profound background in mathematics and physics, they are the coefficients of Fourier transform of images or shapes, containing all the information of them and can be used to reconstruct or recover images or shapes themselves. This property was formulated as the fundamental theorem \cite{Hu1962}\cite{Flusser2017}.

\subsection{Moment invariants and classification}
The research of moment invariants and their applications in image processing was
first proposed in \cite{Hu1962}, where seven moment invariants in 2D were given.
Those are normalized moment invariants and keep invariant under the transformations
of translation, scale and rotation Eq.~\eqref{eq:4}.

\begin{equation}
\begin{aligned}
\label{eq:4}
I_1 &= \mu_{20} + \mu_{02} \\
I_2 &= (\mu_{20} - \mu_{02})^2 + 4 \mu_{11}^2 \\
I_3 &= (\mu_{30} - 3\mu_{12})^2 + (3\mu_{21} - \mu_{03})^2 \\
I_4 &= (\mu_{30} + \mu_{12})^2 + (\mu_{21} + \mu_{03})^2 \\
I_5 &= (\mu_{30} - 3\mu_{12})(\mu_{30} + \mu_{12})[(\mu_{30} + \mu_{12})^2 \\
& \ \ \ \ - 3(\mu_{21} + \mu_{03})^2] + (3\mu_{21} - \mu_{03})(\mu_{21} + \mu_{03}) \\
& \ \ \ \ * [3(\mu_{30} + \mu_{12})^2 - (\mu_{21} + \mu_{03})^2] \\
I_6 &= (\mu_{20} - \mu_{02})[(\mu_{30} + \mu_{12})^2 - (\mu_{21} + \mu_{03})^2] \\
& \ \ \ \ + 4\mu_{11}(\mu_{30} + \mu_{12})(\mu_{21} + \mu_{03}) \\
I_7 &= (3\mu_{21} - \mu_{03})(\mu_{30} + \mu_{12})[(\mu_{30} + \mu_{12})^2 \\
& \ \ \ \ - 3(\mu_{21} + \mu_{03})^2] -(\mu_{30} - 3 \mu_{12})(\mu_{21} + \mu_{03}) \\
& \ \ \ \ *[3(\mu_{30} + \mu_{12})^2 - (\mu_{21} + \mu_{03})^2]
\end{aligned}
\end{equation}

Geometrically, there are two kinds of isometries: one is identical or congruent; the
other is reflection or chiral. Till now in research of moment invariants, not much
attention has been given to the second category, although reflection transform is an
independent transform group. Hu's seven invariants are usually regarded as invariants
for translation, scale and rotation. An interesting fact is that Hu's invariants include
both kinds of invariants, which all belong to isometric invariants in the strict
geometrical meaning. One type of moment invariants (e.g. the seventh in Eq.~\eqref{eq:4}) was
called skew invariant as Hu named it, which could be useful in distinguishing 'mirror
images', for it changed its sign under 'improper rotation', that is reflection or mirror.
Although Hu gave two types of moment invariants, it was ignored and misused in
some thereafter practices, which may partially explain why the applications ofmoment invariants in computer vision and object recognition, have not been as
satisfied as expected.

In this paper, the kind of skew invariants is called reflection invariants. The other
invariants that remain the same under reflection transformation or improper rotation
are called isometry invariants. Such a classification of moment invariants serves better
for symmetry detection.

The multiple integrals of invariant cores proposed by \cite{Xu2008} provide a general
way to build any moment invariants in any dimensions or manifolds. This technique
has been further reformulated as two generating functions by \cite{Li2017}

\begin{figure}[thb]
  \centering
  \includegraphics[width=0.79\linewidth]{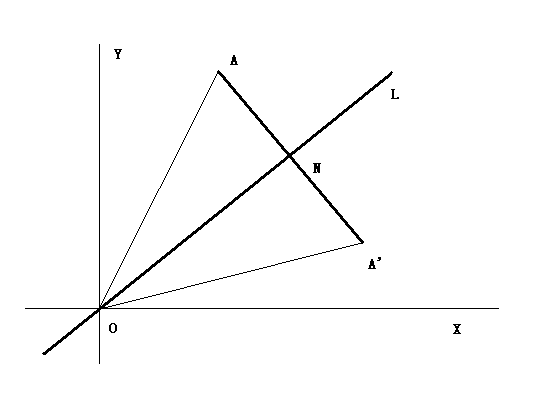}
  \caption{\label{fig:fig2}
  The point $A$ and $A'$ are symmetric with respect to line $L$, where line $N$ is perpendicular
to $L$.  }
\end{figure}

To illustrate the different structures of two kinds of invariants and complete the
classification, first consider a line L in 2D passing through the origin. Let line N
perpendicular to L, as shown in Figure~\ref{fig:fig2}. Suppose the equation of $L: ax + by = 0$,
with $a^2 + b^2 = 1$.

The reflection about line L can be expressed by Householder transform (also known
as Householder reflection) \cite{Householder} , which is a linear transformation with the matrix $H_2$:

\begin{equation}
\label{eq:5}
H_2 = I - 2 V\cdot V^T = det \left|
\begin{array}{cc}
1 - 2a^2 &  -2ab \\
\\
-2ab     &  1-2b^2 \\
\end{array} \right|
\end{equation}

where I is the unit matrix, $V^T = [a b], a^2 + b^2 = 1$, and the determinant of $H_2$ is $-1$.

With reflection transform, moment invariants can be classified. If their signs are
unchanged under an arbitrary reflection transform expressed in Eq.~\eqref{eq:5}, they are isometry
invariants,otherwise they are reflection invariants.

Take the first invariant $I_1 = \mu_{20} + \mu_{02} $ in Eq.~\eqref{eq:4} as an example, according to Householder matrix Eq.~\eqref{eq:5}, the transformation is

\begin{equation}
\begin{aligned}
\label{eq:6}
x^{'} &= (1 - 2a^2)x - 2aby \\
y^{'} &= -2abx + (1-2b^2)y \\
\end{aligned}
\end{equation}

\begin{figure}[thb]
  \centering
  \includegraphics[width=0.79\linewidth]{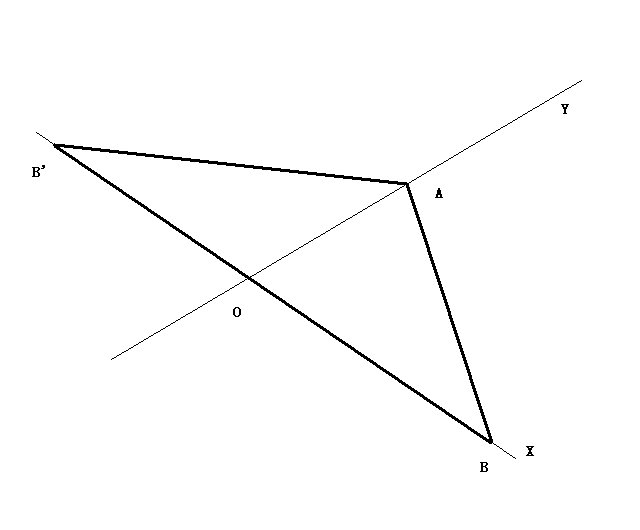}
  \includegraphics[width=0.79\linewidth]{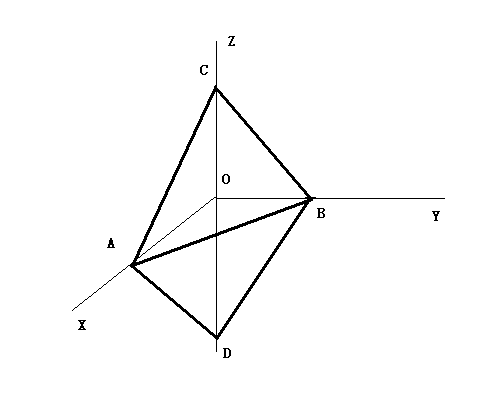}
  \caption{\label{fig:fig3}
   \textbf{Top:} An example in 2D, where triangle $AOB'$ and $AOB$ are reflection-symmetric with respect to line $AO$, with $OB=OB'$. The coordinates are $A(0,3)$, $B(3 \sqrt{3},0)$, $B'(-3 \sqrt{3},0)$, $O(0,0)$.
	\textbf{Bottom:}  An example in 3D, where tetrahedron $AOBD$ and $AOBC$ are reflection-symmetric
with respect to plane $AOB$, with $OC=OD$. The coordinates are $A(0,0,1)$, $B(\sqrt{3},0,0)$, $C(0,2,0)$, $D(0,-2,0)$, $O(0,0,0)$. }
\end{figure}

Considering Eq.~\eqref{eq:1}, it is easy to verify that transformed $I_{1}^{'} = I_1$. The same result holds for
$I_2 ~ I_6$ in Eq.~\eqref{eq:4}, they show no change under the reflection transformation and are
all isometric invariants. The last one $I_7$, which changes its sign, is therefore a
reflection invariant. This coincides with Hu's observation, where he called it as skew
invariant. See Figure~\ref{fig:fig3} and Table~\ref{tbl:tab1} for the result of numerical calculation.

\begin{table}[thb]
  \centering
  \caption{\label{tbl:tab1}
  The seven 2D moment invariants of two reflection triangles. }
  \centering
  \includegraphics[width=0.99\linewidth]{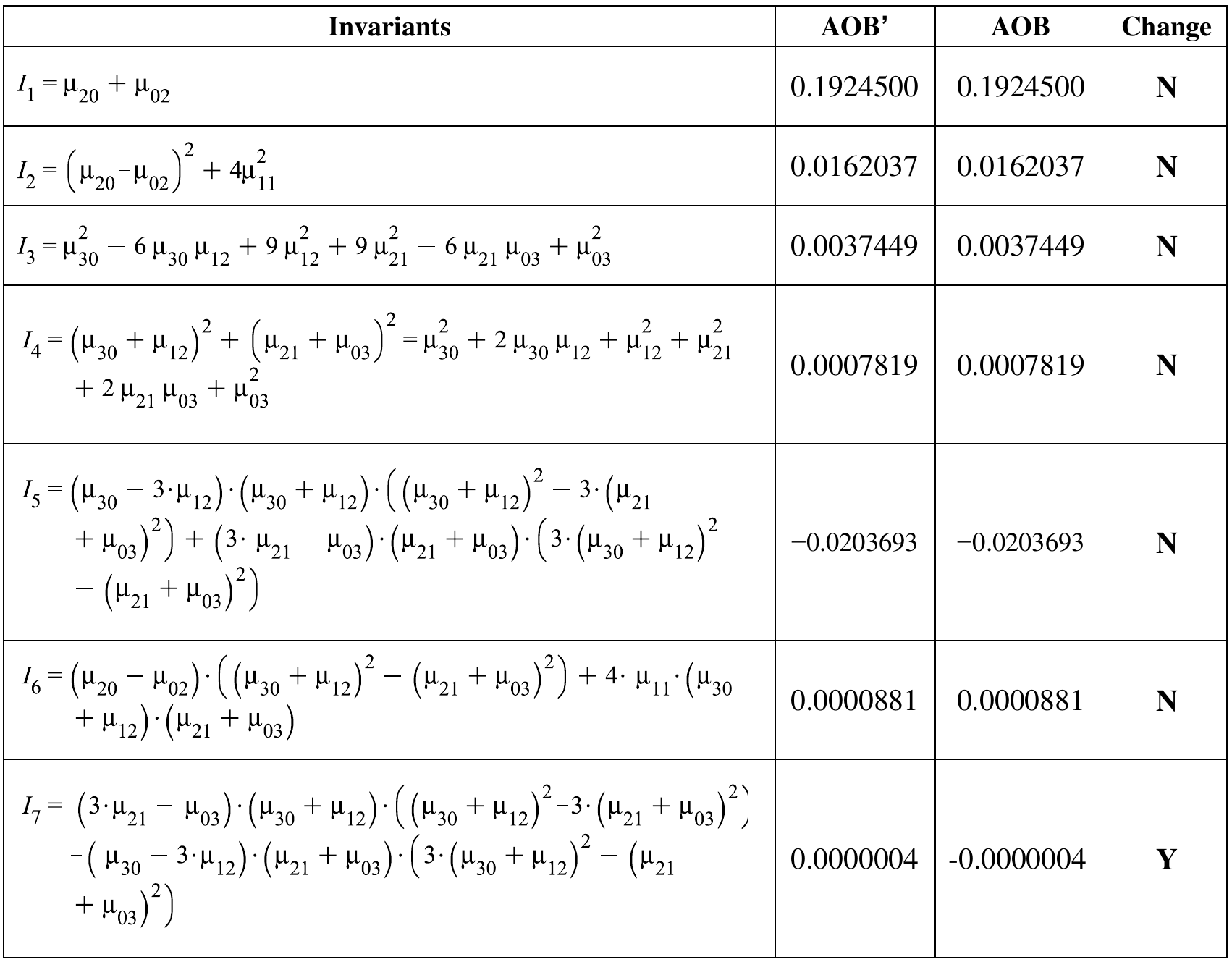}
\end{table}

\subsection{ Reflection and isometric invariants in 2D and 3D}

The classification of invariants based on reflection provides a new perspective for
shape analysis, based one which, two extensions are important. One is to find more
reflection invariants in 2D, and the other is to extend the invariants to 3D space.

According to the technique given by \cite{Xu2008} and the method above, we have
constructed and tested 2D moment invariants within order 3 degree 3. Theoretically,
there may be more in high orders or higher degrees, as Flusser showed some in order
four \cite{Flusser2000} . But the higher order or degree invariants cause more complexity in
calculations and unstable in practical uses.

Two new reflection moment invariants we have discovered are

\begin{equation}
\begin{aligned}
\label{eq:7}
(\mu_{20} - \mu_{02})&(\mu_{03}\mu_{12} + 2\mu_{12}\mu_{21} + \mu_{21}\mu_{30}) \\
&+ \mu_{11}(\mu_{03}^2 + \mu_{12}^2 - \mu_{21}^2 - \mu_{30}^2) \\
\\
(\mu_{20} - \mu_{02})& (\mu_{03}\mu_{30} - \mu_{12}\mu_{21}) \\
&+ 2\mu_{11}(\mu_{21}(\mu_{03} + \mu_{21}) - \mu_{12}(\mu_{12} + \mu_{30})) \\
\end{aligned}
\end{equation}

Similarly, Householder transformation matrix $H_3$ in 3D is defined as following:

\begin{equation}
\label{eq:8}
H_3 = I - 2 V\cdot V^T = det \left|
\begin{array}{ccc}
1 - 2a^2 &  -2ab   &  -2ac \\
\\
-2ab     &  1-2b^2 &  -2bc \\
\\
-2ac     &  -2bc   &  1-2c^2 \\
\end{array} \right|
\end{equation}

And there are some existing moment invariants in 3D \cite{Xu2008} \cite{Lo1989}:

\begin{equation}
\begin{aligned}
\label{eq:9}
I_1 &= \mu_{200} + \mu_{020} + \mu_{002} \\
I_2 &= \mu_{200}\mu_{020} + \mu_{200}\mu_{002} + \mu_{020}\mu_{002} - \mu_{110}^2 \\
& \ \ \ \ - \mu_{101}^2 - \mu_{011}^2 \\
I_3 &= \mu_{200}\mu_{020}\mu_{002} + 2\mu_{110}\mu_{101}\mu_{011} - \mu_{002}\mu_{110}^2 \\
& \ \ \ \ - \mu_{020}\mu_{101}^2  - \mu_{002}\mu_{011}^2 \\
\end{aligned}
\end{equation}

To find a reflection moment invariant like in 2D, we have checked all known 3D
moment invariants, and there are no reflection moment invariants, all of which are
isometric invariants.

See Table~\ref{tbl:tab2} for the result of numerical calculation.

\begin{table}[thb]
  \caption{\label{tbl:tab2}
   The three 3D invariants of two reflection tetrahedra. }
  \centering
  \includegraphics[width=0.99\linewidth]{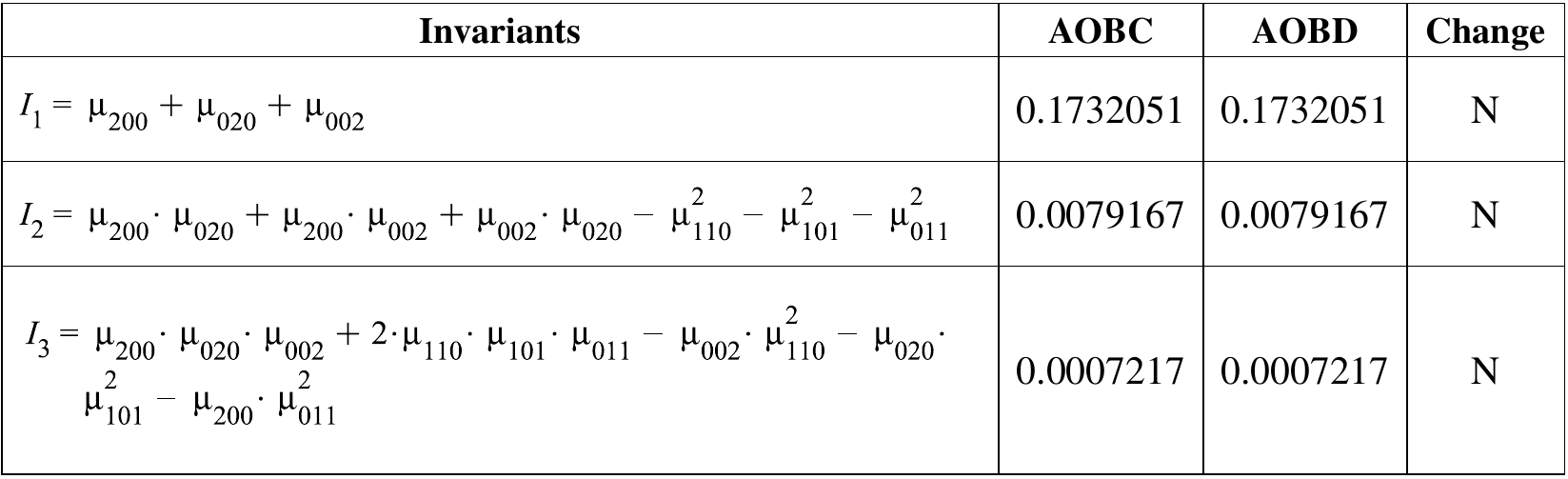}
\end{table}

\subsection{Structure of reflection invariants}
\cite{Li2017} has reformulated the geometric cores as two simple generating functions,
which can be used to define invariants. The generating functions of well-known Hu's
seven invariants are as following:

\begin{equation}
\begin{aligned}
\label{eq:10}
I_1 &\Leftrightarrow f(1,1) \\ 
I_2 &\Leftrightarrow (f(1,2))^2 - 2(g(1,2))^2 \\
I_3 &\Leftrightarrow (f(1,2))^3 - 3(g(1,2))^2f(1,2) \\
I_4 &\Leftrightarrow f(1,2)f(1,1)f(2,2) \\
I_5 &\Leftrightarrow f(2,2)f(3,3)f(4,4)[f(2,1)f(3,1)f(4,1) \\
& \ \ \ \ -f(2,1)g(3, 1)g(4,1)-g(2,1)g(3,1)f(4,1) \\
& \ \ \ \ -g(2,1)f(3,1)g(4,1)] \\
I_6 &\Leftrightarrow f(2,2)f(3,3)[f(1,2)f(1,3) - g(1,2)g(1,3)] \\
I_7 &\Leftrightarrow f(2,2)f(3,3)f(4,4)[g(2,1)f(3,1)f(4,1) \\
& \ \ \ \ -g(2,1)g(3,1)g(4,1)+f(2,1)g(3,1)f(4,1) \\
& \ \ \ \ +f(2,1)f(3,1)g(4,1)]
\end{aligned}
\end{equation}

where $f(i,j)$ and $g(i,j)$ are showing bellow

\begin{equation}
\begin{aligned}
\label{eq:11}
f(i,j) &= (x_i, y_i)(x_j, y_j) = x_{i}x_{j} + y_{i}y_{j} \\
\\
g(i,j) &= det \left|
\begin{array}{cc}
x_i &  y_i \\
x_j &  y_j \\
\end{array} \right|
= x_{i}y_{j} - y_{j}y_{i}\\
\end{aligned}
\end{equation}

Since the invariants are multiple integrals shown in Eq.~\eqref{eq:1}, the first invariant in
Eq.~\eqref{eq:10} in the form of generating functions in Eq.~\eqref{eq:12} can be proved equivalent as in
Eq.~\eqref{eq:4}

\begin{equation}
\begin{aligned}
\label{eq:12}
&\int \int f(1,1)\rho(x, y)dxdy \\
&= \int \int (x_{1}x_{1} + y_{1}y_{1}) \rho(x_1, y_1)dx_{1}dy_1 \\
&= \int \int x_{1}^2 \rho(x_1, y_1)dx_{1}dy_1 + \int \int y_{1}^2 \rho(x_1, y_1)dx_{1}dy_1 \\
&= \mu_{20} + \mu_{02} \\
&= I_1
\end{aligned}
\end{equation}

Where $\rho(x,y)$ is the density function of the image.

Through observation, Hu's seven invariants are not the simplest format, they are
polynomials and can be further split into simpler forms. They consist of multiplication
of generating functions $f$ and $g$, which were called primitive invariants (PIs) or
ShapeDNA \cite{Li2017}.

As a result, Hu's seven invariants are composed of sixteen PIs, with the seventh
contains four. Under calculations, those the four PIs are identical.

Since the seventh invariants is a reflection invariant, it should possess some intrinsic
property that differentiate from the first six invariants. We have the following
conclusion based on observation:

\textbf{Proposition 1:} A necessary and sufficient condition for a PI to be a reflection
invariant is that it is composed of odd number of $g$ functions.

This can be explained by the properties of two generating functions. Function $f$ is in
the form of inner product, which will not change sign under reflection transformations;
then the sign of the invariants depends on function $g$ only. When there are odd
number of $g$ functions in a PI, the composed invariants will change its sign under
reflection transformations.

We have found the generating functions of two reflection invariants in Eq.~\eqref{eq:7}:

\begin{equation}
\begin{aligned}
\label{eq:13}
s_1 &= g(1,2)f(1,2)f(2,3)f(3,3) \\
s_2 &= f(1,2)g(1,3)g(2,3)g(2,3) \\
\end{aligned}
\end{equation}

It is noticed that \cite{Flusser2000} suggested using six order three invariants, two of them are
reflection invariants, including Hu's seventh. But the combination of reflection
invariants with isometric invariants is not encouraged, since they can cause problems
with the reflection transformation.


Similarly, we have found two reflection invariants in 3D, which are shown below in the form of generating functions, and the form of moments can be obtained without difficulty.

\begin{equation}
\begin{aligned}
\label{eq:14}
S_1 &= G(1,2,3)F(1,1)F(1,3)F(2,3) \\
S_2 &= G(1,2,3)F(1,1)F(1,2)F(3,3) \\
\end{aligned}
\end{equation}

\begin{equation}
\begin{aligned}
\label{eq:15}
F(i,j) &= (x_i, y_i, z_i)(x_j, y_j, z_j) = x_{i}x_{j} + y_{i}y_{j} + z_{i}z_{j} \\
\\
G(i,j, k) &= det \left|
\begin{array}{ccc}
x_i &  y_i & z_i \\
x_j &  y_j & z_j \\
x_k &  y_k & z_k \\
\end{array} \right| \\
&= x_{i}y_{j}z_{k} + x_{j}y_{k}z_{i} + x_{k}y_{i}z_{j} \\
&- x_{i}y_{k}z_{j} - x_{j}y_{i}z_{k} - x_{k}y_{j}z_{i} \\
\end{aligned}
\end{equation}

where $G(i,j,k)$ is the determinant of matrix with three vectors $P_m(x_m, y_m, z_m)$;
$F(i,j)$ is scalar product of two vectors. And the calculation of Eq.~\eqref{eq:14} is straight
forward \cite{Xu2008},\cite{Li2017}.

Based on the properties of reflection invariants, we have

\textbf{Proposition 2:} If none of reflection invariants is equal to zero, then there is no symmetry.

\section{ Directional moment and symmetry detection}
\label{sec:3}

The detection of symmetry is critical for shape analysis. Usually it is a tedious and
time consuming task. There are many algorithms to directly detect symmetry, but
none of them avoid complexity. Shen et al proposed a generalized complex moment
to 2D images \cite{Shen1999} , and Martinet et al defined a generalized moment to 3D shapes \cite{Martinet2006}.
Both are close-form solutions and can accurately detect the symmetry. However, the
former can only be used in 2D and the latter is a little complicated to solve in 3D,
which involves an expression of trigonometric function in terms of spherical
harmonics.

We have the following observation:

\textbf{Proposition 3:} In 2D, if two shapes are reflection-symmetric with respect to a common axis, then the
summation of their projections onto the normal line of the axis is zero.

\begin{figure}[thb]
  \centering
  \includegraphics[width=0.79\linewidth]{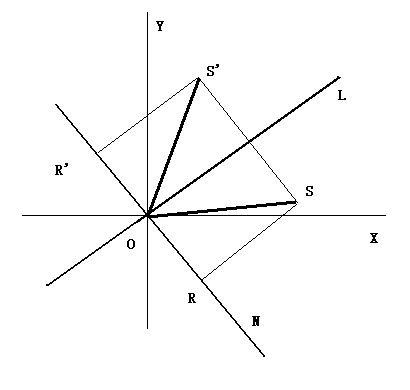}
  \caption{\label{fig:fig4}
   If 2D vector $S$ and $S'$ are reflection-symmetric about the line $L$, then the sum of their
projections $R$ and $R'$ on the normal line $N$ of $L$ is zero. }
\end{figure}

The situation is illustrated in Figure~\ref{fig:fig4}.

Similarly, we have following propositions:

\textbf{Proposition 4:} In 3D, if two shapes are reflection-symmetric with respect to a common plane, then
the summation of their projections onto the normal line of the plane is zero.

\textbf{Proposition 5:} In 2D, if a shape is even rotation-symmetric about its centroid, then the summation of
its projections onto any line passing through the centroid is zero.

\textbf{Proposition 6:} In 3D, if a shape is even rotation-symmetric about an axis passing through its centroid,
then the summation of its projections onto any line passing through the centroid and
perpendicular to the axis is zero.

The above observations are all related to vector projection and can be expressed as
scalar or dot product of vectors within a framework of integral. Therefore, the
definition of directional moment in 3D space shows bellow.

\textbf{Definition 3:} For a shape $S$ defined in a closed volume $V$ in $R^3$ , its order $k$ directional moment (DM) is referred to as

\begin{equation}
\label{eq:16}
M^k(\varphi,\theta) = \int_v (r \cdot d)^k \rho(x,y,z)dv
\end{equation}

where $r = r(x, y, z)$ is the point vector with the origin at the centroid of the shape; $d = d(sin(\varphi)cos(\theta), sin(\varphi)sin(\theta), cos(\varphi))$ is the unit direction vector in spherical coordinates, $ 0 < \varphi < \pi, 0 < \theta < 2\pi, k=1,2, \cdots, n; \rho = \rho(x,y,z)$ is the density distribution function; $\cdot$ is the dot product of vectors.

The geometric meaning of the definition is the summation of projections of point
vectors along the given direction. And the DM expressed in this way fits all kinds of
shapes, including area and boundary line in 2D, spatial curves, curved surfaces, and
volumes in 3D etc.

For a given order $k$, the DM in Eq.~\eqref{eq:16} can be expanded and expressed as in Eq.~\eqref{eq:1}. For
example, when $k=2$, the $M^2(\varphi,\theta)$ is

\begin{equation}
\begin{aligned}
\label{eq:17}
M^2(\varphi,\theta) &= \int (r \cdot d)^2 \rho(x,y,z)dv \\
&= \int(x sin(\varphi)cos(\theta) + y sin(\varphi)sin(\theta) \\
&+ z cos(\varphi))^2 \rho(x,y,z)dv \\
\end{aligned}
\end{equation}

Considering Eq.~\eqref{eq:1},

\begin{equation}
\begin{aligned}
\label{eq:18}
&M^2(\varphi,\theta) = \mu_{200}sin(\varphi)^2cos(\theta)^2 \\
&+ 2\mu_{110}sin(\varphi)^2cos(\theta)sin(\theta) \\
&+ 2\mu_{101}sin(\varphi)cos(\theta)cos(\varphi) + \mu_{020}sin(\varphi)^2sin(\theta)^2 \\
&+ 2\mu_{011}sin(\varphi)sin(\theta)cos(\varphi) + \mu_{002}cos(\varphi)^2 \\
\end{aligned}
\end{equation}

For a fixed order $k$, the DM is a function with two parameters of the direction. If DM
is zero in one direction, then there may be a normal line aligned in that direction,
perpendicular to a reflection axis in 2D, or to a reflection plane in 3D. The complexity
of symmetry detection is now reduced to find all the solutions of $M^k_{\theta}=0$ or $M^k(\varphi,
\theta)=0$ respectively, which is a system of trigonometric equations.

For the definition of DM in 3D, when $k$ is chosen, there is only one equation for two
parameters. Since the number of equations is less than the number of unknown
parameters, it cannot be solved completely. If $k$ is an even number, the DM is the
even-power summation of all projections. Obviously, there exist extremes (minimum)
for the equation. Its two first-order partial derivatives can construct a working system,
to which the solution should contain the candidates of the normal directions of
possible reflection planes. When $k$ is a larger number, more roots of the equation
could be found.

For example, let $k=2$, two first-order derivatives of Eq.~\eqref{eq:18} are in below.

\begin{equation}
\begin{aligned}
\label{eq:19}
&\mu_{200}sin(\varphi)cos(\theta)^2cos(\varphi) \\
&+ 2\mu_{110}sin(\varphi)cos(\theta)sin(\theta)cos(\varphi) \\
&+ \mu_{101}cos(\varphi)^2cos(\theta) - \mu_{101}sin(\varphi)^2cos(\theta) \\
&+ \mu_{020}sin(\varphi)sin(\theta)^2cos(\varphi) + \mu_{011}cos(\varphi)^2sin(\theta) \\
&- \mu_{011}sin(\varphi)^2sin(\theta) - \mu_{002}cos(\varphi)sin(\varphi) \\
&= 0 \\
\\
&-\mu_{200}sin(\varphi)^2cos(\theta)sin(\theta) - \mu_{110}sin(\varphi)^2sin(\theta)^2 \\
&+ \mu_{110}sin(\varphi)^2cos(\theta)^2 - \mu_{101}sin(\varphi)sin(\theta)cos(\varphi) \\
&+ \mu_{020}sin(\varphi)^2sin(\theta)cos(\theta) + \mu_{011}sin(\varphi)cos(\theta)cos(\varphi) \\
&= 0 \\
\end{aligned}
\end{equation}

The DM is an extension of traditional moment, which plays a central role in the algorithms to solve symmetry detection problems. This is an extension or complement of method given in \cite{Martinet2006} and \cite{Shen1999}, and unifies the solution for different dimensions.

\section{Algorithms and experiments}
\label{sec:4}

The combination of the directional moment and the reflection invariant provides a
powerful tool for the solution of reflection symmetry directly. And it can be used to
deal with rotation symmetry as well, with the help of isometric invariants.

The algorithm for symmetry detection and discrimination in 2D is described as
following:

\noindent\rule{8cm}{0.4pt} \\
\textbf{Algorithm 1: 2D shape symmetry axes of reflection and even-rotation } \\
\noindent\rule{8cm}{0.4pt} \\
\begin{algorithmic}
\STATE For a given shape $S$ in 2D Cartesian coordinates
\STATE \textbf{1.} calculate the moment $M_{00}$, $M_{10}$, and $M_{01}$ by Eq.~\eqref{eq:1} in the given coordinates
\STATE \textbf{2.} find the centroid by $x^{'} = M_{10} / M_{00}$, $y^{'} = M_{01} / M_{00}$
\STATE \textbf{3.} (virtually) translate the shape, with the centroid at the origin
\STATE \textbf{4.} calculate the central moment $\mu_{ij}$ to proper order
\STATE \textbf{5.} quickly determine if exists any reflection symmetry axis
\IF{ none of reflection invariants in Eq.~\eqref{eq:7} is equal to zero}
\STATE \textbf{stop}
\ENDIF
\STATE \textbf{6.} construct DM according to Eq.~\eqref{eq:16}
\STATE \textbf{7.} determine a non-constant DM $M^k(\theta)$ with minimum order $k$
\STATE \textbf{8.} solve $M^k(\theta) = 0$ to get all roots ${\theta_i}$, $i=1,\cdots, n$ as candidates
\FOR{ $N=1, 2,\cdots, n$}
\STATE test $\theta_N$ by treating it as normal line of the to-be reflection axis:
\STATE \textbf{a.} divide the shape into two parts $P_L, P_R$ along the axis
\STATE \textbf{b.} calculate the reflection invariants $RI_{L,j}$ and $RI_{R,j}$, $j=1,2$ according to Eq.~\eqref{eq:7}
\IF{$RI_{L,j} = -RI_{R,j}$, $j=1,2$}
\STATE the axis is a reflection direction
\ENDIF
\STATE \textbf{c.} calculate isometric invariants $II_{j,m}$, $j=L,R$ for $P_L$, $P_R$, $m = 1,2,\cdots,6$ for $I_m$ according to Eq.~\eqref{eq:4}
\IF{$II_{L,m} = II_{R,m}$, $m = 1,2,\cdots,6$}
\STATE the axis is an even rotation direction
\ELSE
								\IF{ $N < n$}
								\STATE $N = N+1$
								\STATE \textbf{continue}
								\ELSE
								\STATE \textbf{break}
								\ENDIF
\ENDIF
\ENDFOR
\STATE \textbf{9.}	check the fold number of even rotation symmetry, if any.
\end{algorithmic}
\noindent\rule{8cm}{0.4pt} \\

Notice that rotation and reflection symmetry may co-exist for some shapes, it is
necessary to detect both of them at the same time.

Similarly, there is an algorithm for 3D.

\noindent\rule{8cm}{0.4pt} \\
\textbf{Algorithm 2:  3D shape symmetry planes of reflection and axes of even-rotation } \\
\noindent\rule{8cm}{0.4pt} \\
\begin{algorithmic}
\STATE For a given shape $S$ in 3D Cartesian coordinates
\STATE \textbf{1.} calculate the moment $M_{000}$, $M_{100}$, $M_{010}$ and $M_{001}$ in the given coordinates
\STATE \textbf{2.} find the centroid by $x^{'} = M_{100} / M_{000}$, $y^{'} = M_{010} / M_{000}$, $z^{'} = M_{001} / M_{000}$
\STATE \textbf{3.} (virtually) translate the shape, with the centroid at the origin
\STATE \textbf{4.} calculate the central moment $\mu_{ijm}$ to proper order
\STATE \textbf{5.} quickly determine if exists any reflection symmetry plane
\IF{ none of reflection invariants in Eq.~\eqref{eq:14} is equal to zero}
\STATE \textbf{stop}
\ENDIF
\STATE \textbf{6.} construct DM according to Eq.~\eqref{eq:16}
\STATE \textbf{7.} determine a non-constant DM $M^k(\varphi,\theta)$ with minimum order $k = 2m$, $m = 1,2,\cdots $
\STATE \textbf{8.} find the derivatives $\partial M^k (\varphi,\theta)/ \partial \varphi$ and $\partial M^k (\varphi,\theta)/ \partial \theta $
\STATE \textbf{9.} solve $\partial M^k (\varphi,\theta)/ \partial \varphi = 0$ and $\partial M^k (\varphi,\theta)/ \partial \theta = 0$ to get all roots ${\varphi_i, \theta_i}$, $i=1, 2, \cdots, n$ as candidates

\FOR{$N=1, 2,\cdots, n$}
\STATE test $(\varphi_N,\theta_N)$ by treating it as normal line of a potential reflection axis:
\STATE \textbf{a.} divide the shape into two parts $P_L$, $P_R$ by the plane passing through the centroid, and taking the line as normal
\STATE \textbf{b.} calculate the reflection invariants $RI_{L,j}$ and $RI_{R,j}$, $j=1,2$ of $P_L, P_R$ according to Eq.~\eqref{eq:14}
\IF{$RI_{L,j} = -RI_{R,j}$, $j=1,2$}
\STATE the plane is a reflection symmetry plane
\ENDIF
\STATE test $(\varphi_N,\theta_N)$ by treating it as axis of even rotation:
\STATE \textbf{c.} divide the shape into two parts $P_L, P_R$ by the plane passing through the centroid and along the line
\STATE \textbf{d.} calculate isometric invariants $II_{j,m}$, $j=L,R$ for $P_L$, $P_R$, $m = 1,2,3$ for $I_m$ according to Eq.~\eqref{eq:9}
\IF{$II_{L,m} = II_{R,m}$, $m = 1,2,3$}
\STATE the axis is an axis of even rotation symmetry
\ELSE
								\IF{ $N < n$}
								\STATE $N = N+1$
								\STATE \textbf{continue}
								\ELSE
								\STATE \textbf{break}
								\ENDIF
\ENDIF
\ENDFOR
\STATE \textbf{10.}	check the fold number of even rotation symmetry, if any.
\end{algorithmic}
\noindent\rule{8cm}{0.4pt} \\

Notice that for rotation symmetry, there are two cases: odd and even symmetry. Based on the two algorithms, the axes of even rotation symmetry can be solved. For pure odd symmetry (non-reflection), things are a little different. 

The first step is to find the potential rotation axis, which is always possible. For a shape in 2D, the axis is the line passing through the centroid and perpendicular to the plane the shape is lying on. For a shape in 3D, it is a line passing through the centroid.

\begin{figure}[thb]
  \centering
  \includegraphics[width=0.79\linewidth]{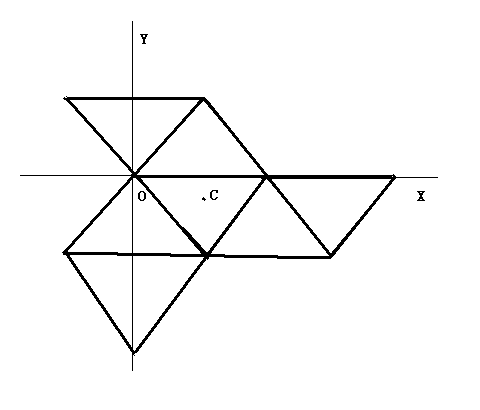}
  \caption{\label{fig:fig5}
   An example of pure odd rotation, where the seven triangles are regular with edge length
1 and the point $C(1/2, - \sqrt{3} /6)$ is the centroid. There exists no symmetry axis on the plane. }
\end{figure}

As an example, a shape derived from \cite{Weyl1952} is shown in Figure~\ref{fig:fig5}. It is composed of
seven regular triangles and there is three-fold rotation symmetry about its centroid
without any reflection symmetry. For its $M^2(\theta)$ is not constant, it is used to solve the
possible symmetry axes. But no real root is found. While using the shape as bounding
faces at the top and the bottom to construct a cylinder, the solution of $M^2(\varphi,\theta)$ gives
the symmetry axis in the third dimension, say $Z$ coordinate.

From the property of rotation symmetry and the definition of DM, if a shape is n-fold rotation symmetry about an axis, then all the n DMs of the shape along n directions with iso-angle distribution on the plane normal to the axis are also symmetric. And all the DMs have the same magnitude. 

Suppose all the axes of rotation symmetry are given, we have algorithm 3 and 4 for the symmetry of rotation for 2D and 3D shapes.

To test the algorithms, we have done some experiments in 2D and 3D respectively,
using Maple 11 as equation solver.

First, an equilateral or regular triangle and a square in 2D are chosen. For the
equilateral triangle, there are three extremes of $M^3(\theta)$, the solution proves the
existence of three normal lines, with respect to one symmetric axis each. As for the
square, all four symmetry axes are obtained from $M^4(\theta)$, for its lower order DMs are
constant.

The five Plato's objects in 3D are selected for the experiment, because they typically
have more symmetries and easy to check, See Table~\ref{tbl:tab3}. Only even orders of DM are
useful in 3D, whose derivatives are employed to find extremes of DMs with two
parameters. For a tetrahedron, there are six reflection symmetry planes, which are
drawn from $M^4(\varphi,\theta)$, for $M^2(\varphi,\theta)$ is constant. There are nine reflection symmetry
planes for a cube and an octahedron, both of which are found from $M^4(\varphi,\theta)$, with
$M^2(\varphi,\theta)$ constant. There are fifteen reflection symmetry planes for a dodecahedron
and an icosahedron, which are obtained from $M^6(\varphi,\theta)$ with $M^2(\varphi,\theta)$ and $M^4(\varphi,\theta)$ being
constant. It is shown that a cube and an octahedron have the same symmetry, the same
to a dodecahedron and an icosahedron, for they are dual entities.

%
%

\noindent\rule{8cm}{0.4pt} \\
\textbf{Algorithm 3: 2D shape symmetry of rotation} \\
\noindent\rule{8cm}{0.4pt} \\
\begin{algorithmic}
\STATE Let $n$, the maximum possible symmetry number
\STATE Determine the first non-constant DM $M^k(\theta)$, $k=2,3,\cdots$

\FOR{$N=2,3,\cdots,n$}
\STATE calculate $m_i = M^k(\theta_i)$, $\theta_i = \theta + \frac{2\pi i}{N}$, $i = 0, 1, \cdots, N-1$
\IF{$m^2_i = m^2_{i+1(mod N)}$, $i = 1, 2, \cdots, N$}
\STATE the shape is N-fold rotation symmetric
\STATE \textbf{break}
\ENDIF
\ENDFOR
\end{algorithmic}
\noindent\rule{8cm}{0.4pt} \\

%

\noindent\rule{8cm}{0.4pt} \\
\textbf{Algorithm 4: 3D shape symmetry of rotation} \\
\noindent\rule{8cm}{0.4pt} \\
\begin{algorithmic}

\FOR{each given possible rotation axis $d_L$, $L=1,2, \cdots, L_{axis}$ axis}
\STATE Determine the first non-constant DM $M^k(\varphi, \theta)$, $k=2,3,\cdots $

\FOR{$N=2,3,\cdots,n$}
\STATE find directions $(\varphi_i,\theta_i)$, $i=1,2, \cdots, N$ with iso-angle-distribution on the plane perpendicular to direction $d_L$
\STATE calculate $m_i = M^k(\varphi_i, \theta_i)$, $i = 1,2, \cdots, N$
\IF{$m^2_i = m^2_{i+1(mod N)}$, $i = 1, 2, \cdots, N$}
\STATE the shape is N-fold rotation symmetric  about axis $d_L$
\STATE \textbf{break}
\ENDIF
\ENDFOR
\ENDFOR
\end{algorithmic}
\noindent\rule{8cm}{0.4pt} \\

Therefore, the reflection symmetry of all Plato's objects is solved by DMs with order
no more than six.

\begin{table}[thb]
  \centering
  \caption{\label{tbl:tab3}
   Reflection symmetry planes for Platonic objects. }
  \centering
  \includegraphics[width=0.99\linewidth]{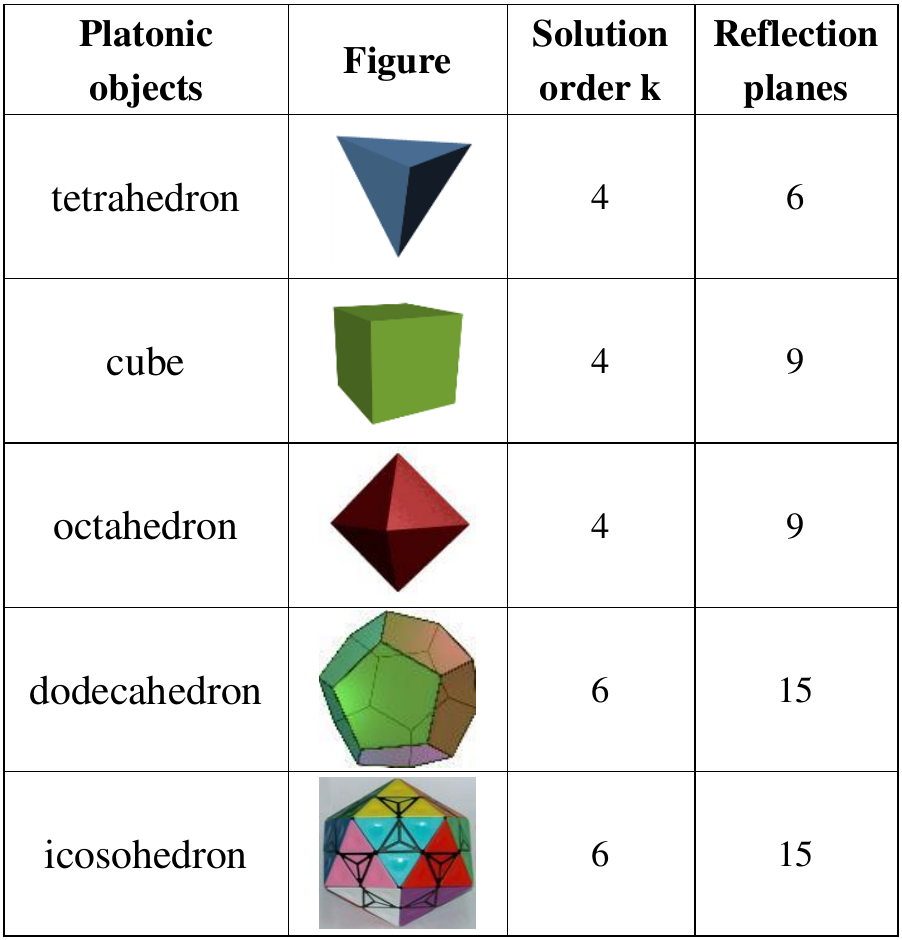}
\end{table}

\section{Discussions}
\label{sec:5}

This paper provides the details of directional moments and reflection invariants,
which provide an easier and more effective way for symmetry detection and
discrimination.

\textbf{(1) The reflection invariant is useful.}

The reflection invariant is the group invariant, since reflection or mirror
transformation is an independent transform group, including translation, scale and
rotation (TSR). There have been many discussions on invariants of TSR, not much on
reflection. In many research areas, chiralty is important, where reflection invariant can
play a key role. The description of reflection invariant in Eq.~\eqref{eq:14} is much simpler than the
form given in \cite{Hattne2011}. If a shape possesses some reflective symmetry, its reflection
invariants are all equal to zero, no matter it is a curve, image, surface, or volume,
Such property can be used in fast evaluating before the symmetry detection and
calculation.

\textbf{(2) The DM is simple to calculate.}

Theoretically, the DM can also be defined in a form of cross or hybrid product of vectors like the definition of generalized moment by \cite{Martinet2006}. We discarded that method, for the summation of projections along one direction is essential and can be expressed by the dot product and determinant precisely. All three forms of vector products result in similar system of equations with the same solution set, but the other two vector products involve higher order and higher degree moments, bringing more complicated expressions, and thus more difficult to solve.

\textbf{(3) The solution of detection and discrimination is optimal.}

Searching for symmetry axes using sophisticated algorithms is unnecessary. And the application of moment invariants is free of comparison between correspondent points, as most work done like in \cite{Kazhdan2003},\cite{Zabrodsky1995}. In addition, applying the reflection transformation is also unnecessary in the context of moment invariants. Referring to the detection of symmetry, the traditional way is to flip twice (one for the original query shapes, and the other for the compared shapes under reflection transformation). In our method, we only need to calculate the moment invariants for flipped shape instead of flipping it, which will be much time efficiency. 

\textbf{(4) The algorithm is robust.}

From the experiments, when there are more symmetry planes or axes, the degree $k$ of DM has to be increased to acquire thorough solutions, and the first non-constant DM gives the solution for reflection symmetry. 

Symmetry detection using DM performs well, with little interference by noises, because it defines a continuous measure of shape symmetry by integrals in a whole instead of partial symmetry \cite{Mitra2006}, \cite{Tagliasacchi2009}. And the directional moment proposed unifies the solution of symmetry for different dimensions and manifolds. The generalized moment by \cite{Martinet2006} is only for surfaces.

\section{Conclusions}
In this paper we have shown that the detection of reflective symmetry can be done in a simple way by solving a trigonometric system derived from DMs, and symmetry discrimination can be achieved by applying the reflection moment invariants. Current experiments show that all the reflection lines or planes can be deterministically found with DMs up to order six. And rotation symmetry of the shape can also be determined similarly. 

In practical uses, DMs and moment invariants are promising to solve symmetry related problems, which plays a key role in shape analysis, as in protein structure, medical care, mechanical design, visual perception, robot vision, architecture, and art etc. The work in this paper provides a useful and powerful tool to the solutions.


%


\section*{Acknowledgment}

This work was partly funded by National Natural Science Foundation of China (Grant No.60573154, 61227802 and 61379082).

\ifCLASSOPTIONcaptionsoff
  \newpage
\fi


\bibliographystyle{IEEEtran}
\bibliography{symmetry}
\end{document}